# Mean Field Theory for Sigmoid Belief Networks

**Lawrence K. Saul**                                                    LKSAUL@PSYCHE.MIT.EDU
**Tommi Jaakkola**                                                        TOMMI@PSYCHE.MIT.EDU
**Michael I. Jordan**                                                    JORDAN@PSYCHE.MIT.EDU
*Center for Biological and Computational Learning*
*Massachusetts Institute of Technology*
*79 Amherst Street, E10-243*
*Cambridge, MA 02139*

## Abstract

We develop a mean field theory for sigmoid belief networks based on ideas from statistical mechanics. Our mean field theory provides a tractable approximation to the true probability distribution in these networks; it also yields a lower bound on the likelihood of evidence. We demonstrate the utility of this framework on a benchmark problem in statistical pattern recognition—the classification of handwritten digits.

## 1. Introduction

Bayesian belief networks (Pearl, 1988; Lauritzen & Spiegelhalter, 1988) provide a rich graphical representation of probabilistic models. The nodes in these networks represent random variables, while the links represent causal influences. These associations endow directed acyclic graphs (DAGs) with a precise probabilistic semantics. The ease of interpretation afforded by this semantics explains the growing appeal of belief networks, now widely used as models of planning, reasoning, and uncertainty.

Inference and learning in belief networks are possible insofar as one can efficiently compute (or approximate) the likelihood of observed patterns of evidence (Buntine, 1994; Russell, Binder, Koller, & Kanazawa, 1995). There exist provably efficient algorithms for computing likelihoods in belief networks with tree or chain-like architectures. In practice, these algorithms also tend to perform well on more general sparse networks. However, for networks in which nodes have many parents, the exact algorithms are too slow (Jensen, Kong, & Kjaefulff, 1995). Indeed, in large networks with dense or layered connectivity, exact methods are intractable as they require summing over an exponentially large number of hidden states.

One approach to dealing with such networks has been to use Gibbs sampling (Pearl, 1988), a stochastic simulation methodology with roots in statistical mechanics (Geman & Geman, 1984). Our approach in this paper relies on a different tool from statistical mechanics—namely, mean field theory (Parisi, 1988). The mean field approximation is well known for probabilistic models that can be represented as undirected graphs—so-called Markov networks. For example, in Boltzmann machines (Ackley, Hinton, & Sejnowski, 1985), mean field learning rules have been shown to yield tremendous savings in time and computation over sampling-based methods (Peterson & Anderson, 1987).

The main motivation for this work was to extend the mean field approximation for undirected graphical models to their directed counterparts. Since belief networks can be transformed to Markov networks, and mean field theories for Markov networks are well known, it is natural to ask why a new framework is required at all. The reason is that probabilistic models which have compact representations as DAGs may have unwieldy representations as undirected graphs. As we shall see, avoiding this complexity and working directly on DAGs requires an extension of existing methods.

In this paper we focus on sigmoid belief networks (Neal, 1992), for which the resulting mean field theory is most straightforward. These are networks of binary random variables whose local conditional distributions are based on log-linear models. We develop a mean field approximation for these networks and use it to compute a lower bound on the likelihood of evidence. Our method applies to arbitrary partial instantiations of the variables in these networks and makes no restrictions on the network topology. Note that once a lower bound is available, a learning procedure can maximize the lower bound; this is useful when the true likelihood itself cannot be computed efficiently. A similar approximation for models of continous random variables is discussed by Jaakkola et al (1995).

The idea of bounding the likelihood in sigmoid belief networks was introduced in a related architecture known as the Helmholtz machine (Hinton, Dayan, Frey, & Neal 1995). A fundamental advance of this work was to establish a framework for approximation that is especially conducive to learning the parameters of layered belief networks. The close connection between this idea and the mean field approximation from statistical mechanics, however, was not developed.

In this paper we hope not only to elucidate this connection, but also to convey a sense of which approximations are likely to generate useful lower bounds while, at the same time, remaining analytically tractable. We develop here what is perhaps the simplest such approximation for belief networks, noting that more sophisticated methods (Jaakkola & Jordan, 1996a; Saul & Jordan, 1995) are also available. It should be emphasized that approximations of some form are required to handle the multilayer neural networks used in statistical pattern recognition. For these networks, exact algorithms are hopelessly intractable; moreover, Gibbs sampling methods are impractically slow.

The organization of this paper is as follows. Section 2 introduces the problems of inference and learning in sigmoid belief networks. Section 3 contains the main contribution of the paper: a tractable mean field theory. Here we present the mean field approximation for sigmoid belief networks and derive a lower bound on the likelihood of instantiated patterns of evidence. Section 4 looks at a mean field algorithm for learning the parameters of sigmoid belief networks. For this algorithm, we give results on a benchmark problem in pattern recognition—the classification of handwritten digits. Finally, section 5 presents our conclusions, as well as future issues for research.

## 2. Sigmoid Belief Networks

The great virtue of belief networks is that they clearly exhibit the conditional dependencies of the underlying probability model. Consider a belief network defined over binary random variables $S = (S_1, S_2, \ldots, S_N)$. We denote the parents of $S_i$ by $\mathrm{pa}(S_i) \subseteq \{S_1, S_2, \ldots S_{i-1}\}$; this is the smallest set of nodes for which

$$P(S_i | S_1, S_2, \ldots, S_{i-1}) = P(S_i | \mathrm{pa}(S_i)). \tag{1}$$

In sigmoid belief networks (Neal, 1992), the conditional distributions attached to each node are based on log-linear models. In particular, the probability that the $i$th node is activated is given by

$$P(S_i = 1 | \mathrm{pa}(S_i)) = \sigma \left( \sum_j J_{ij} S_j + h_i \right), \tag{2}$$

where $J_{ij}$ and $h_i$ are the weights and biases in the network, and

$$\sigma(z) = \frac{1}{1 + e^{-z}} \tag{3}$$

is the sigmoid function shown in Figure 1. In sigmoid belief networks, we have $J_{ij} = 0$ for $S_j \notin \mathrm{pa}(S_i)$; moreover, $J_{ij} = 0$ for $j \geq i$ since the network's structure is that of a directed acyclic graph.

The sigmoid function in eq. (2) provides a compact parametrization of the conditional probability distributions[1] in eq. (2) used to propagate beliefs. In particular, $P(S_i | \mathrm{pa}(S_i))$ depends on $\mathrm{pa}(S_i)$ only through a sum of weighted inputs, where the weights may be viewed as the parameters in a

---

1. The relation to noisy-OR models is discussed in appendix A.

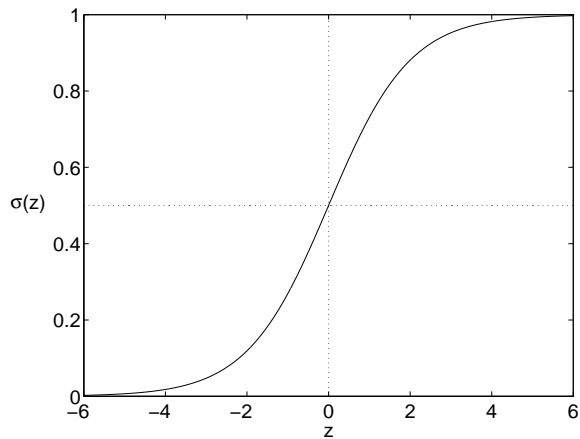

Figure 1: Sigmoid function $\sigma(z) = [1 + e^{-z}]^{-1}$. If $z$ is the sum of weighted inputs to node $S$, then $P(S = 1|z) = \sigma(z)$ is the conditional probability that node $S$ is activated.

logistic regression (McCullagh & Nelder, 1983). The conditional probability distribution for $S_i$ may be summarized as:

$$P(S_i|\text{pa}(S_i)) = \frac{\exp\left[\left(\sum_j J_{ij}S_j + h_i\right)S_i\right]}{1 + \exp\left[\sum_j J_{ij}S_j + h_i\right]}.$$ (4)

Note that substituting $S_i = 1$ in eq. (4) recovers the result in eq. (2). Combining eqs. (1) and (4), we may write the joint probability distribution over the variables in the network as:

$$P(S) = \prod_i P(S_i|\text{pa}(S_i))$$ (5)

$$= \prod_i \left\{ \frac{\exp\left[\left(\sum_j J_{ij}S_j + h_i\right)S_i\right]}{1 + \exp\left[\sum_j J_{ij}S_j + h_i\right]} \right\}.$$ (6)

The denominator in eq. (6) ensures that the probability distribution is normalized to unity.

We now turn to the problem of inference in sigmoid belief networks. Absorbing evidence divides the units in the belief network into two types, visible and hidden. The visible units (or "evidence nodes") are those for which we have instantiated values; the hidden units are those for which we do not. When there is no possible ambiguity, we will use $H$ and $V$ to denote the subsets of hidden and visible units. Using Bayes' rule, inference is done under the conditional distribution

$$P(H|V) = \frac{P(H, V)}{P(V)},$$ (7)

where

$$P(V) = \sum_H P(H, V)$$ (8)

is the likelihood of the evidence $V$. In principle, the likelihood may be computed by summing over all $2^{|H|}$ configurations of the hidden units. Unfortunately, this calculation is intractable in large, densely connected networks. This intractability presents a major obstacle to learning parameters for these networks, as nearly all procedures for statistical estimation require frequent estimates of the likelihood. The calculations for exact probabilistic inference are beset by the same difficulties.

Unable to compute $P(V)$ or work directly with $P(H|V)$, we will resort to an approximation from statistical physics known as mean field theory.

## 3. Mean Field Theory

The mean field approximation appears under a multitude of guises in the physics literature; indeed, it is "almost as old as statistical mechanics" (Itzykson & Drouffe, 1991). Let us briefly explain how it acquired its name and why it is so ubiquitous. In the physical models described by Markov networks, the variables $S_i$ represent localized magnetic moments (e.g., at the sites of a crystal lattice), and the sums $\sum_j J_{ij} S_j + h_i$ represent local magnetic fields. Roughly speaking, in certain cases a central limit theorem may be applied to these sums, and a useful approximation is to ignore the fluctuations in these fields and replace them by their mean value—hence the name, "mean field" theory. In some models, this is an excellent approximation; in others, a poor one. Because of its simplicity, however, it is widely used as a first step in understanding many types of physical phenomena.

Though this explains the philological origins of mean field theory, there are in fact many ways to derive what amounts to the same approximation (Parisi, 1988). In this paper we present the formulation most appropriate for inference and learning in graphical models. In particular, we view mean field theory as a principled method for approximating an intractable graphical model by a tractable one. This is done via a variational principle that chooses the parameters of the tractable model to minimize an entropic measure of error.

The basic framework of mean field theory remains the same for directed graphs, though we have found it necessary to introduce extra mean field parameters in addition to the usual ones. As in Markov networks, one finds a set of nonlinear equations for the mean field parameters that can be solved by iteration. In practice, we have found this iteration to converge fairly quickly and to scale well to large networks.

Let us now return to the problem posed at the end of the last section. There we found that for many belief networks, it was intractable to decompose the joint distribution as $P(S) = P(H|V)P(V)$, where $P(V)$ was the likelihood of the evidence $V$. For the purposes of probabilistic modeling, mean field theory has two main virtues. First, it provides a tractable approximation, $Q(H|V) \approx P(H|V)$, to the conditional distributions required for inference. Second, it provides a lower bound on the likelihoods required for learning.

Let us first consider the origin of the lower bound. Clearly, for any approximating distribution $Q(H|V)$, we have the equality:

$$\ln P(V) \;=\; \ln \sum_H P(H, V) \tag{9}$$

$$=\; \ln \sum_H Q(H|V) \cdot \left[\frac{P(H, V)}{Q(H|V)}\right]. \tag{10}$$

To obtain a lower bound, we now apply Jensen's inequality (Cover & Thomas, 1991), pushing the logarithm through the sum over hidden states and into the expectation:

$$\ln P(V) \geq \sum_H Q(H|V) \ln \left[\frac{P(H, V)}{Q(H|V)}\right]. \tag{11}$$

It is straightforward to verify that the difference between the left and right hand side of eq. (11) is the Kullback-Leibler divergence (Cover & Thomas, 1991):

$$\mathrm{KL}(Q\|P) = \sum_H Q(H|V) \ln \left[\frac{Q(H|V)}{P(H|V)}\right]. \tag{12}$$

Thus, the better the approximation to $P(H|V)$, the tighter the bound on $\ln P(V)$.

Anticipating the connection to statistical mechanics, we will refer to $Q(H|V)$ as the mean field distribution. It is natural to divide the calculation of the bound into two components, both of which are particular averages over this approximating distribution. These components are the mean field entropy and energy; the overall bound is given by their difference:

$$\ln P(V) \geq \left( -\sum_H Q(H|V) \ln Q(H|V) \right) - \left( -\sum_H Q(H|V) \ln P(H,V) \right). \tag{13}$$

Both terms have physical interpretations. The first measures the amount of uncertainty in the mean-field distribution and follows the standard definition of entropy. The second measures the average value[2] of $-\ln P(H,V)$; the name "energy" arises from interpreting the probability distributions in belief networks as Boltzmann distributions[3] at unit temperature. In this case, the energy of each network configuration is given (up to a constant) by minus the logarithm of its probability under the Boltzmann distribution. In sigmoid belief networks, the energy has the form

$$-\ln P(H,V) = -\sum_{ij} J_{ij} S_i S_j - \sum_i h_i S_i + \sum_i \ln \left[ 1 + \exp \left( \sum_j J_{ij} S_j + h_i \right) \right], \tag{14}$$

as follows from eq. (6). The first two terms in this equation are familiar from Markov networks with pairwise interactions (Hertz, Krogh, & Palmer, 1991); the last term is peculiar to sigmoid belief networks. Note that the overall energy is neither a linear function of the weights nor a polynomial function of the units. This is the price we pay in sigmoid belief networks for identifying $P(H|V)$ as a Boltzmann distribution and the log-likelihood $P(V)$ as its partition function. Note that this identification was made implicitly in the form of eqs. (7) and (8).

The bound in eq. (11) is valid for any probability distribution $Q(H|V)$. To make use of it, however, we must choose a distribution that enables us to evaluate the right hand side of eq. (11). Consider the factorized distribution

$$Q(H|V) = \prod_{i \in H} \mu_i^{S_i} (1 - \mu_i)^{1 - S_i}, \tag{15}$$

in which the binary hidden units $\{S_i\}_{i \in H}$ appear as independent Bernoulli variables with adjustable means $\mu_i$. A mean field approximation is obtained by substituting the factorized distribution, eq. (15), for the true Boltzmann distribution, eq. (7). It may seem that this approximation replaces the rich probabilistic dependencies in $P(H|V)$ by an impoverished assumption of complete factorizability. Though this is true to some degree, the reader should keep in mind that the values we choose for $\{\mu_i\}_{i \in H}$ (and hence the statistics of the hidden units) will depend on the evidence $V$.

The best approximation of the form, eq. (15), is found by choosing the mean values, $\{\mu_i\}_{i \in H}$, that minimize the Kullback-Leibler divergence, $KL(Q||P)$. This is equivalent to minimizing the gap between the true log-likelihood, $\ln P(V)$, and the lower bound obtained from mean field theory. The

---

2. A similar average is performed in the E-step of an EM algorithm (Dempster, Laird, & Rubin, 1977); the difference here is that the average is performed over the mean field distribution, $Q(H|V)$, rather than the true posterior, $P(H|V)$. For a related discussion, see Neal & Hinton (1993).

3. Our terminology is as follows. Let $S$ denote the degrees of freedom in a statistical mechanical system. The *energy* of the system, $\mathcal{E}(S)$, is a real-valued function of these degrees of freedom, and the *Boltzmann distribution*

$$P(S) = \frac{e^{-\beta \mathcal{E}(S)}}{\sum_S e^{-\beta \mathcal{E}(S)}}$$

defines a probability distribution over the possible configurations of $S$. The parameter $\beta$ is the *inverse temperature*; it serves to calibrate the energy scale and will be fixed to unity in our discussion of belief networks. Finally, the sum in the denominator—known as the *partition function*—ensures that the Boltzmann distribution is normalized to unity.

mean field bound on the log-likelihood may be calculated by substituting eq. (15) into the right hand side of eq. (11). The result of this calculation is

$$\ln P(V) \geq \sum_{ij} J_{ij}\mu_i\mu_j + \sum_i h_i\mu_i - \sum_i \left\langle \ln\left[1 + e^{\sum_j J_{ij}S_j + h_i}\right]\right\rangle \qquad (16)$$
$$- \sum_i \left[\mu_i \ln\mu_i + (1-\mu_i)\ln(1-\mu_i)\right],$$

where $\langle\cdot\rangle$ indicates an expectation value over the mean field distribution, eq. (15). The terms in the first line of eq. (16) represent the mean field energy, derived from eq. (14); those in the second represent the mean field entropy. In a slight abuse of notation, we have defined mean values $\mu_i$ for the visible units; these of course are set to the instantiated values $\mu_i \in \{0, 1\}$.

Note that to compute the average energy in the mean field approximation, we must find the expected value of $\langle\ln[1 + e^{z_i}]\rangle$, where $z_i = \sum_j J_{ij}S_j + h_i$ is the sum of weighted inputs to the $i$th unit in the belief network. Unfortunately, even under the mean field assumption that the hidden units are uncorrelated, this average does not have a simple closed form. This term does not arise in the mean field theory for Markov networks with pairwise interactions; again, it is peculiar to sigmoid belief networks.

In principal, the average may be performed by enumerating the possible states of $\mathrm{pa}(S_i)$. The result of this calculation, however, would be an extremely unwieldy function of the parameters in the belief network. This reflects the fact that in general, the sigmoid belief network defined by the weights $J_{ij}$ has an equivalent Markov network with $N$th order interactions and not pairwise ones. To avoid this complexity, we must develop a mean field theory that works directly on DAGs.

How we handle the expected value of $\langle\ln[1 + e^{z_i}]\rangle$ is what distinguishes our mean field theory from previous ones. Unable to compute this term exactly, we resort to another bound. Note that for any random variable $z$ and any real number $\xi$, we have the equality:

$$\langle\ln[1 + e^z]\rangle = \left\langle \ln\left[e^{\xi z}e^{-\xi z}(1 + e^z)\right]\right\rangle \qquad (17)$$
$$= \xi\langle z\rangle + \left\langle\ln[e^{-\xi z} + e^{(1-\xi)z}]\right\rangle. \qquad (18)$$

We can upper bound the right hand side by applying Jensen's inequality in the opposite direction as before, pulling the logarithm outside the expectation:

$$\langle\ln[1 + e^z]\rangle \leq \xi\langle z\rangle + \ln\left\langle e^{-\xi z} + e^{(1-\xi)z}\right\rangle. \qquad (19)$$

Setting $\xi = 0$ in eq. (19) gives the standard bound: $\langle\ln(1 + e^z)\rangle \leq \ln\langle 1 + e^z\rangle$. A tighter bound (Seung, 1995) can be obtained, however, by allowing non-zero values of $\xi$. This is illustrated in Figure 2 for the special case where $z$ is a Gaussian distributed random variable with zero mean and unit variance. The bound in eq. (19) has two useful properties which we state here without proof: (i) the right hand side is a convex function of $\xi$; (ii) the value of $\xi$ which minimizes this function occurs in the interval $\xi \in [0, 1]$. Thus, provided it is possible to evaluate eq. (19) for different values of $\xi$, the tightest bound of this form can be found by a simple one-dimensional minimization.

The above bound can be put to immediate use by attaching an extra mean field parameter $\xi_i$ to each unit in the belief network. We can then upper bound the intractable terms in the mean field energy by

$$\left\langle \ln\left[1 + e^{\sum_j J_{ij}S_j + h_i}\right]\right\rangle \leq \xi_i\left(\sum_j J_{ij}\mu_j + h_i\right) + \ln\left\langle e^{-\xi_i z_i} + e^{(1-\xi_i)z_i}\right\rangle, \qquad (20)$$

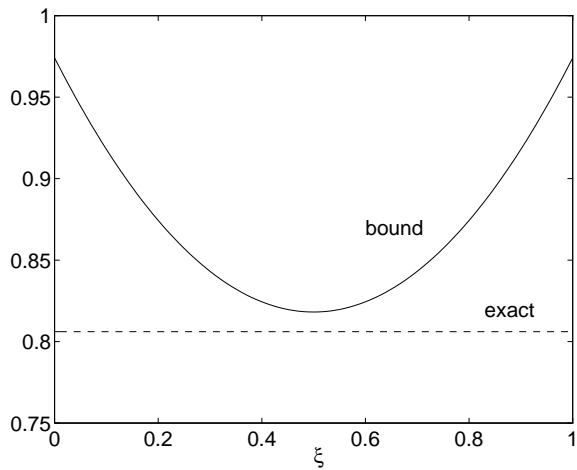

Figure 2: Bound in eq. (19) for the case where $z$ is normally distributed with zero mean and unit variance. In this case, the exact result is $\langle \ln(1 + e^z) \rangle = 0.806$; the bound gives $\min_\xi \left\{ \ln[e^{\frac{1}{2}\xi^2} + e^{\frac{1}{2}(1-\xi)^2}] \right\} = 0.818$. The standard bound from Jensen's inequality occurs at $\xi = 0$ and gives $0.974$.

where $z_i = \sum_j J_{ij} S_j + h_i$. The expectations inside the logarithm can be evaluated exactly for the factorial distribution, eq. (15); for example,

$$\langle e^{-\xi_i z_i} \rangle = e^{-\xi_i h_i} \prod_j \left( 1 - \mu_j + \mu_j e^{-\xi_i J_{ij}} \right). \tag{21}$$

A similar result holds for $\langle e^{(1-\xi_i)z_i} \rangle$. Though these averages are tractable, we will tend not to write them out in what follows. The reader, however, should keep in mind that these averages do not present any difficulty; they are simply averages over products of independent random variables, as opposed to sums.

Assembling the terms in eqs. (16) and (20) gives a lower bound $\ln P(V) \geq \mathcal{L}_V$,

$$
\begin{aligned}
\mathcal{L}_V = & \sum_{ij} J_{ij} \mu_i \mu_j + \sum_i h_i \mu_i - \sum_i \xi_i \left( \sum_j J_{ij} \mu_j + h_i \right) \\
& - \sum_i \ln \left\langle e^{-\xi_i z_i} + e^{(1-\xi_i)z_i} \right\rangle + \sum_i \left[ \mu_i \ln \mu_i + (1 - \mu_i) \ln(1 - \mu_i) \right],
\end{aligned}
\tag{22}
$$

on the log-likelihood of the evidence $V$. So far we have not specified the parameters $\{\mu_i\}_{i \in H}$ and $\{\xi_i\}$; in particular, the bound in eq. (22) is valid for any choice of parameters. We naturally seek the values that maximize the right hand side of eq. (22). Suppose we fix the mean values $\{\mu_i\}_{i \in H}$ and ask for the parameters $\{\xi_i\}$ that yield the tightest possible bound. Note that the right hand side of eq. (22) does not couple terms with $\xi_i$ that belong to different units in the network. The minimization over $\{\xi_i\}$ therefore reduces to $N$ independent minimizations over the interval $[0, 1]$. These can be done by any number of standard methods (Press, Flannery, Teukolsky, & Vetterling, 1986).

To choose the means, we set the gradients of the bound with respect to $\{\mu_i\}_{i \in H}$ equal to zero. To this end, let us define the intermediate matrix:

$$K_{ij} = -\frac{\partial}{\partial \mu_j} \ln \left\langle e^{-\xi_i z_i} + e^{(1-\xi_i)z_i} \right\rangle, \tag{23}$$

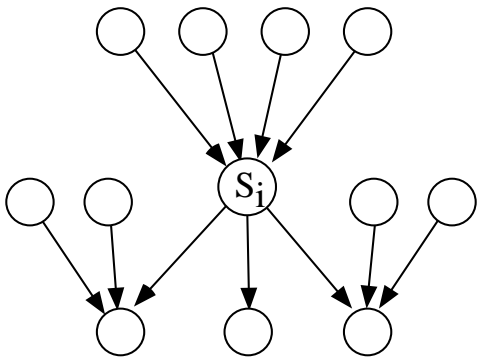

Figure 3: The Markov blanket of unit $S_i$ includes its parents and children, as well as the other parents of its children.

where $z_i$ is the weighted sum of inputs to $i$th unit. Note that $K_{ij}$ is zero unless $S_j$ is a parent of $S_i$; in other words, it has the same connectivity as the weight matrix $J_{ij}$. Within the mean field approximation, $K_{ij}$ measures the parental influence of $S_j$ on $S_i$ given the instantiated evidence $V$. The degree of correlation (positive or negative) is measured relative to the other parents of $S_i$.

The matrix elements of $K$ may be evaluated by expanding the expectations as in eq. (21); a full derivation is given in appendix B. Setting the gradient $\partial \mathcal{L}_V / \partial \mu_i$ equal to zero gives the final mean field equation:

$$\mu_i = \sigma \left( h_i + \sum_j \left[ J_{ij} \mu_j + J_{ji}(\mu_j - \xi_j) + K_{ji} \right] \right), \tag{24}$$

where $\sigma(\cdot)$ is the sigmoid function. The argument of the sigmoid function may be viewed as an effective input to the $i$th unit in the belief network. This effective input is composed of terms from the unit's Markov blanket (Pearl, 1988), shown in Figure 3; in particular, these terms take into account the unit's internal bias, the values of its parents and children, and, through the matrix $K_{ji}$, the values of its children's other parents. In solving these equations by iteration, the values of the instantiated units are propagated throughout the entire network. An analogous propagation of information occurs in exact algorithms (Lauritzen & Spiegelhalter, 1988) to compute likelihoods in belief networks.

While the factorized approximation to the true posterior is not exact, the mean field equations set the parameters $\{\mu_i\}_{i \in H}$ to values which make the approximation as accurate as possible. This in turn translates into the tightest mean field bound on the log-likelihood. The overall procedure for bounding the log-likelihood thus consists of two alternating steps: (i) update $\{\xi_i\}$ for fixed $\{\mu_i\}$; (ii) update $\{\mu_i\}_{i \in H}$ for fixed $\{\xi_i\}$. The first step involves $N$ independent minimizations over the interval $[0, 1]$; the second is done by iterating the mean field equations. In practice, the steps are repeated until the mean field bound on the log-likelihood converges[4] to a desired degree of accuracy.

The quality of the bound depends on two approximations: the complete factorizability of the mean field distribution, eq. (15), and the logarithm bound, eq. (19). How reliable are these approximations in belief networks? To study this question, we performed numerical experiments on the three layer belief network shown in Figure 4. The advantage of working with such a small network (2x4x6) is that true likelihoods can be computed by exact enumeration. We considered the particular event that all the units in the bottom layer were instantiated to zero. For this event, we compared the mean field bound on the likelihood to its true value, obtained by enumerating the states in the top two layers. This was done for 10000 random networks whose weights and biases were uniformly distributed between -1 and 1. Figure 5 (left) shows the histogram of the relative error in log likelihood, computed as $\mathcal{L}_V / \ln P(V) - 1$; for these networks, the mean relative error is 1.6%. Figure 5 (right) shows the histogram that results from assuming that all states in the bottom layer occur with equal probability; in this case the relative error was computed as $(\ln 2^{-6}) / \ln P(V) - 1$. For this "uniform" approximation, the root mean square relative error is 22.6%. The large discrepancy between these results suggests that mean field theory can provide a useful lower bound on the likelihood in certain belief networks. Of course, what ultimately matters is the behavior of mean field theory in networks that solve meaningful problems. This is the subject of the next section.

---

4. It can be shown that asynchronous updates of the mean field parameters lead to monotonic increases in the lower bound (just as in the case of Markov networks).

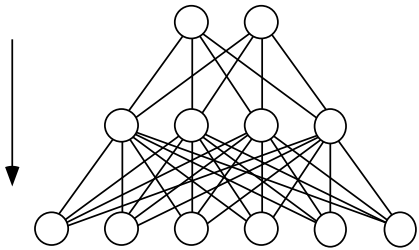

Figure 4: Three layer belief network (2x4x6) with top-down propagation of beliefs. To model the images of handwritten digits in section 4, we used 8x24x64 networks where units in the bottom layer encoded pixel values in 8x8 bitmaps.

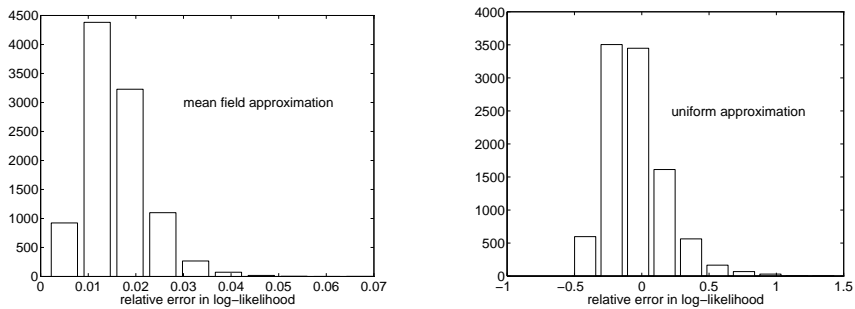

Figure 5: Histograms of relative error in log-likelihood over 10000 randomly generated three layer networks. At left: the relative error from the mean field approximation; at right: the relative error if all states in the bottom layer are assumed to occur with equal probability. The log-likelihood was computed for the event that the all the nodes in the bottom layer were instantiated to zero.

## 4. Learning

One attractive use of sigmoid belief networks is to perform density estimation in high dimensional input spaces. This is a problem in parameter estimation: given a set of patterns over particular units in the belief network, find the set of weights $J_{ij}$ and biases $h_i$ that assign high probability to these patterns. Clearly, the ability to compute likelihoods lies at the crux of any algorithm for learning the parameters in belief networks.

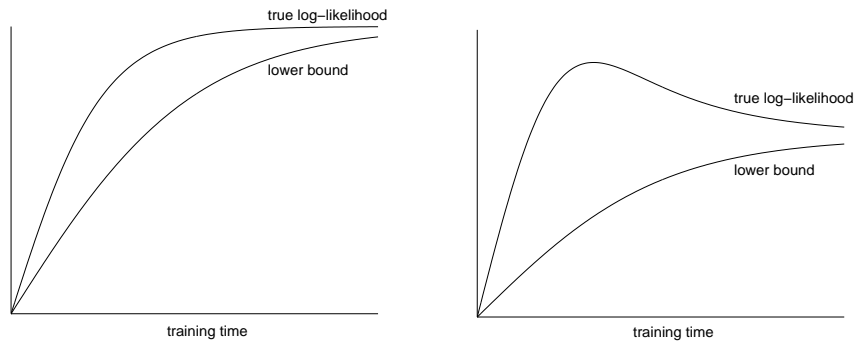

Figure 6: Relationship between the true log-likelihood and its lower bound during learning. One possibility (at left) is that both increase together. The other is that the true log-likelihood decreases, closing the gap between itself and the bound. The latter can be viewed as a form of regularization.

Mean field algorithms provide a strategy for discovering appropriate values of $J_{ij}$ and $h_i$ without resort to Gibbs sampling. Consider, for instance, the following procedure. For each pattern in the training set, solve the mean field equations for $\{\mu_i, \xi_i\}$ and compute the associated bound on the log-likelihood, $\mathcal{L}_V$. Next, adapt the weights in the belief network by gradient ascent[5] in the mean field bound,

$$\Delta J_{ij} = \eta \frac{\partial \mathcal{L}_V}{\partial J_{ij}} \tag{25}$$

$$\Delta h_i = \eta \frac{\partial \mathcal{L}_V}{\partial h_i}, \tag{26}$$

where $\eta$ is a suitably chosen learning rate. Finally, cycle through the patterns in the training set, maximizing their likelihoods[6] for a fixed number of iterations or until one detects the onset of overfitting (e.g., by cross-validation).

The above procedure uses a lower bound on the log-likelihood as a cost function for training belief networks (Hinton, Dayan, Frey, & Neal, 1995). The fact that we have a lower bound on the log-likelihood, rather than an upper bound, is of course crucial to the success of this learning algorithm. Adjusting the weights to maximize this lower bound can affect the true log-likelihood in two ways (see Figure 6). Either the true log-likelihood increases, moving in the same direction as the bound, or the true log-likelihood decreases, closing the gap between these two quantities. For the purposes of maximum likelihood estimation, the first outcome is clearly desirable; the second, though less desirable, can also be viewed in a positive light. In this case, the mean field approximation is acting as a regularizer, steering the network toward simple, factorial solutions even at the expense of lower likelihood estimates.

We tested this algorithm by building a maximum-likelihood classifier for images of handwritten digits. The data consisted of 11000 examples of handwritten digits [0-9] compiled by the U.S. Postal Service Office of Advanced Technology. The examples were preprocessed to produce 8x8 binary images, as shown in Figure 7. For each digit, we divided the available data into a training set with 700 examples and a test set with 400 examples. We then trained a three layer network[7] (see

---

5. Expressions for the gradients of $\mathcal{L}_V$ are given in the appendix B.

6. Of course, one can also incorporate prior distributions over the weights and biases and maximize an approximation to the log posterior probability of the training set.

7. There are many possible architectures that could be chosen for the purpose of density estimation; we used layered networks to permit a comparison with previous benchmarks on this data set.

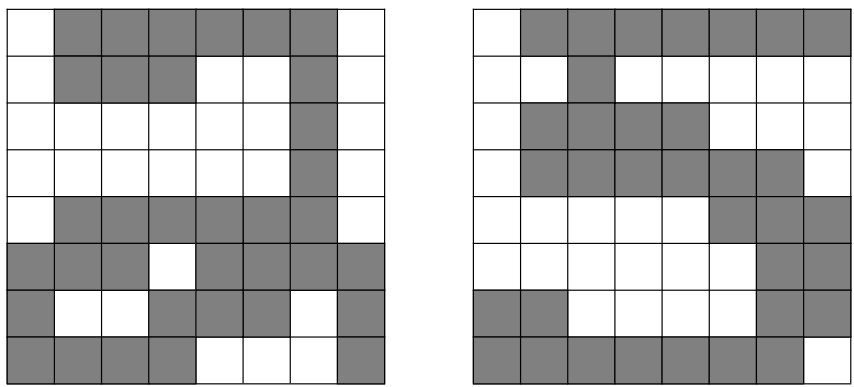

Figure 7: Binary images of handwritten digits: two and five.

|   | 0 | 1 | 2 | 3 | 4 | 5 | 6 | 7 | 8 | 9 |
|---|---|---|---|---|---|---|---|---|---|---|
| 0 | 388 | 2 | 2 | 0 | 1 | 3 | 0 | 0 | 4 | 0 |
| 1 | 0 | 393 | 0 | 0 | 0 | 1 | 0 | 0 | 6 | 0 |
| 2 | 1 | 2 | 376 | 1 | 3 | 0 | 4 | 0 | 13 | 0 |
| 3 | 0 | 2 | 4 | 373 | 0 | 12 | 0 | 0 | 6 | 3 |
| 4 | 0 | 0 | 2 | 0 | 383 | 0 | 1 | 2 | 2 | 10 |
| 5 | 0 | 2 | 1 | 13 | 0 | 377 | 2 | 0 | 4 | 1 |
| 6 | 1 | 4 | 2 | 0 | 1 | 6 | 386 | 0 | 0 | 0 |
| 7 | 0 | 1 | 0 | 0 | 0 | 0 | 0 | 388 | 3 | 8 |
| 8 | 1 | 9 | 1 | 7 | 0 | 7 | 1 | 1 | 369 | 4 |
| 9 | 0 | 4 | 0 | 0 | 0 | 0 | 0 | 8 | 5 | 383 |

Table 1: Confusion matrix for digit classification. The entry in the $i$th row and $j$th column counts the number of times that digit $i$ was classified as digit $j$.

Figure 4) on each digit, sweeping through each training set five times with learning rate $\eta = 0.05$. The networks had 8 units in the top layers, 24 units in the middle layer, and 64 units in the bottom layer, making them far too large to be treated with exact methods.

After training, we classified the digits in each test set by the network that assigned them the highest likelihood. Table 1 shows the confusion matrix in which the $ij$th entry counts the number of times digit $i$ was classified as digit $j$. There were 184 errors in classification (out of a possible 4000), yielding an overall error rate of 4.6%. Table 2 gives the performance of various other algorithms on the same partition of this data set. Table 3 shows the average log-likelihood score of each network on the digits in its test set. (Note that these scores are actually lower bounds.) These scores are normalized so that a network with zero weights and biases (i.e., one in which all 8x8 patterns are equally likely) would receive a score of -1. As expected, digits with relatively simple constructions (e.g., zeros, ones, and sevens) are more easily modeled than the rest.

Both measures of performance—error rate and log-likelihood score—are competitive with previously published results (Hinton, Dayan, Frey, & Neal, 1995) on this data set. The success of the algorithm affirms both the strategy of maximizing a lower bound and the utility of the mean field approximation. Though similar results can be obtained via Gibbs sampling, this seems to require considerably more computation than methods based on maximizing a lower bound (Frey, Dayan, & Hinton, 1995).

| algorithm | classification error |
|---|---|
| nearest neighbor | 6.7% |
| back-propagation | 5.6% |
| wake-sleep | 4.8% |
| mean field | 4.6% |

Table 2: Classification error rates for the data set of handwritten digits. The first three were reported by Hinton et al (1995).

| digit | log-likelihood score |
|---|---|
| 0 | -0.447 |
| 1 | -0.296 |
| 2 | -0.636 |
| 3 | -0.583 |
| 4 | -0.574 |
| 5 | -0.565 |
| 6 | -0.515 |
| 7 | -0.434 |
| 8 | -0.569 |
| 9 | -0.495 |
| all | -0.511 |

Table 3: Normalized log-likelihood score for each network on the digits in its test set. To obtain the raw score, multiply by $400 \times 64 \times \ln 2$. The last row shows the score averaged across all digits.

## 5. Discussion

Endowing networks with probabilistic semantics provides a unified framework for incorporating prior knowledge, handling missing data, and performing inference under uncertainty. Probabilistic calculations, however, can quickly become intractable, so it is important to develop techniques that approximate probability distributions in a flexible manner. This is especially true for networks with multilayer architectures and large numbers of hidden units. Exact algorithms and Gibbs sampling methods are not generally practical for such networks; approximations are required.

In this paper we have developed a mean field approximation for sigmoid belief networks. As a computational tool, our mean field theory has two main virtues: first, it provides a tractable approximation to the conditional distributions required for inference; second, it provides a lower bound on the likelihoods required for learning.

The problem of computing exact likelihoods in belief networks is NP-hard (Cooper, 1990); the same is true for approximating likelihoods to within a guaranteed degree of accuracy (Dagum & Luby, 1993). It follows that one cannot establish universal guarantees for the accuracy of the mean field approximation. For certain networks, clearly, the mean field approximation is bound to fail—it cannot capture logical constraints or strong correlations between fluctuating units. Our preliminary results, however, suggest that these worst-case results do not apply to all belief networks. It is worth noting, moreover, that all the above qualifications apply to Markov networks, and that in this domain, mean field methods are already well-established.

The idea of bounding the likelihood in sigmoid belief networks was introduced in a related architecture known as the Helmholtz machine (Hinton, Dayan, Neal, & Zemel, 1995). The formalism in this paper differs in a number of respects from the Helmholtz machine. Most importantly, it enables one to compute a rigorous lower bound on the likelihood. This cannot be said for the wake-sleep algorithm (Frey, Hinton, & Dayan, 1995), which relies on sampling-based methods, or the heuristic approximation of Dayan et al (1995), which does not guarantee a rigorous lower bound. Also, our mean field theory—which takes the place of the "recognition model" of the Helmholtz machine—applies generally to sigmoid belief networks with or without layered structure. Moreover, it places no restrictions on the locations of visible units; they may occur anywhere within the network—an important feature for handling problems with missing data. Of course, these advantages are not accrued without extra computational demands and more complicated learning rules.

In recent work that builds on the theory presented here, we have begun to relax the assumption of complete factorizability in eq. (15). In general, one would expect more sophisticated approximations to the Boltzmann distribution to yield tighter bounds on the log-likelihood. The challenge here is to find distributions that allow for correlations between hidden units while remaining computationally tractable. By tractable, we mean that the choice of $Q(H|V)$ must enable one to evaluate (or at least upper bound) the right hand side of eq. (13). Extensions of this kind include mixture models (Jaakkola & Jordan, 1996) and/or partially factorized distributions (Saul & Jordan, 1995) that exploit the presence of tractable substructures in the original network. Our approach in this paper has been to work out the simplest mean field theory that is computationally tractable, but clearly better results will be obtained by tailoring the approximation to the problem at hand.

## Appendix A. Sigmoid versus Noisy-OR

The semantics of the sigmoid function are similar, but not identical, to the noisy-OR gates (Pearl, 1988) more commonly found in the belief network literature. Noisy-OR gates use the weights in the network to represent independent causal events. In this case, the probability that unit $S_i$ is activated is given by

$$P(S_i = 1|\text{pa}(S_i)) = 1 - \prod_j (1 - p_{ij})^{S_j} \tag{27}$$

where $p_{ij}$ is the probability that $S_j = 1$ causes $S_i = 1$ in the absence of all other causal events. If we define the weights of a noisy-OR belief network by $\theta_{ij} = -\ln(1 - p_{ij})$, it follows that

$$p(S_i|\text{pa}(S_i)) = \rho \left( \sum_j \theta_{ij} S_j \right), \tag{28}$$

where

$$\rho(z) = 1 - e^{-z} \tag{29}$$

is the noisy-OR gating function. Comparing this to the sigmoid function, eq. (3), we see that both model $P(S_i|\text{pa}(S_i))$ as a monotonically increasing function of a sum of weighted inputs. The main difference is that in noisy-OR networks, the weights $\theta_{ij}$ are constrained to be positive by an underlying set of probabilities, $p_{ij}$. Recently, Jaakkola and Jordan (1996b) have developed a mean field approximation for noisy-OR belief networks.

## Appendix B. Gradients

Here we provide expressions for the gradients that appear in eqs. (23), (25) and (26). As usual, let $z_i = \sum_j J_{ij} S_j + h_i$ denote the sum of inputs into unit $S_i$. Under the factorial distribution, eq. (15), we can compute the averages:

$$\langle e^{-\xi_i z_i} \rangle = e^{-\xi_i h_i} \prod_j \left[ 1 - \mu_j + \mu_j e^{-\xi_i J_{ij}} \right], \tag{30}$$

$$\langle e^{(1-\xi_i) z_i} \rangle = e^{(1-\xi_i) h_i} \prod_j \left[ 1 - \mu_j + \mu_j e^{(1-\xi_i) J_{ij}} \right]. \tag{31}$$

For each unit in the network, let us define the quantity

$$\phi_i = \frac{\langle e^{(1-\xi_i) z_i} \rangle}{\langle e^{-\xi_i z_i} + e^{(1-\xi_i) z_i} \rangle}. \tag{32}$$

Note that $\phi_i$ lies between zero and one. With this definition, we can write the matrix elements in eq. (23) as:

$$K_{ij} = \frac{(1 - \phi_i)(1 - e^{-\xi_i J_{ij}})}{1 - \mu_j + \mu_j e^{-\xi_i J_{ij}}} + \frac{\phi_i (1 - e^{(1-\xi_i) J_{ij}})}{1 - \mu_j + \mu_j e^{(1-\xi_i) J_{ij}}}. \tag{33}$$

The gradients in eqs. (25) and (26) are found by similar means. For the weights, we have

$$\frac{\partial \mathcal{L}_V}{\partial J_{ij}} = -(\xi_i - \mu_i)\mu_j + \frac{(1 - \phi_i)\xi_i \mu_j e^{-\xi_i J_{ij}}}{1 - \mu_j + \mu_j e^{-\xi_i J_{ij}}} - \frac{\phi_i (1 - \xi_i)\mu_j e^{(1-\xi_i) J_{ij}}}{1 - \mu_j + \mu_j e^{(1-\xi_i) J_{ij}}}. \tag{34}$$

Likewise, for the biases, we have

$$\frac{\partial \mathcal{L}_V}{\partial h_i} = \mu_i - \phi_i. \tag{35}$$

Finally, we note that one may obtain simpler gradients at the expense of introducing a weaker bound than eq. (19). This can be advantageous when speed of computation is more important than the quality of the bound. All the experiments in this paper used the bound in eq. (19).

## Acknowledgements

We are especially grateful to P. Dayan, G. Hinton, B. Frey, R. Neal, and H. Seung for sharing early versions of their manuscripts and for providing many stimulating discussions about this work. The paper was also improved greatly by the comments of several anonymous reviewers. To facilitate comparisons with similar methods, the results reported in this paper used images that were preprocessed at the University of Toronto. The authors acknowledge support from NSF grant CDA-9404932, ONR grant N00014-94-1-0777, ATR Research Laboratories, and Siemens Corporation.